\documentclass[letterpaper, 10 pt, conference]{ieeeconf}  % Comment this line out if you need a4paper

\IEEEoverridecommandlockouts                              % This command is only needed if 
\usepackage{graphics} % for pdf, bitmapped graphics files
\usepackage[demo]{graphicx}
\usepackage{subcaption}
\usepackage{xcolor,soul}
\usepackage{epsfig} % for postscript graphics files
\usepackage{mathptmx} % assumes new font selection scheme installed
\usepackage{times} % assumes new font selection scheme installed
\usepackage{amsmath} % assumes amsmath package installed
\usepackage{amssymb}  % assumes amsmath package installed
\usepackage{mathtools}
\usepackage{pgfplots}
\pgfplotsset{compat=1.9}
\usepackage{multirow}

\def\baselinestretch{0.976}

\usepackage{booktabs}
\usepackage{caption}
\newcommand{\figref}[2][{}]{\hyperref[#2]{\figurename~\ref{#2}#1}} 
\newcommand{\tableref}[2][{}]{\hyperref[#2]{\tablename~\ref{#2}#1}} 
\newcommand{\etoe}{\text{end-to-end}}

\newcommand\norm[1]{\left\lVert#1\right\rVert}
\title{\LARGE \bf
Quantity over Quality: Training an AV Motion Planner \\ with Large Scale Commodity Vision Data
}

\author{Lukas Platinsky$^{1}$, Tayyab Naseer$^{1}$, Hui Chen$^{1}$, Ben Haines$^{1}$, Haoyue Zhu$^{1}$, Hugo Grimmett$^{1}$, Luca Del Pero$^{1*}$% <-this % stops a space
\thanks{$^{1}$ All authors affiliated with Woven Planet United Kingdom Limited. Email address:
{\tt\footnotesize first.last@woven-planet.global, luca.d.pero@woven-planet.global(*)}}%
}

\begin{document}

\maketitle
\thispagestyle{empty}
\pagestyle{empty}

%%%%%%%%%%%%%%%%%%%%%%%%%%%%%%%%%%%%%%%%%%%%%%%%%%%%%%%%%%%%%%%%%%%%%%%%%%%%%%%%
\begin{abstract}
With the Autonomous Vehicle (AV) industry shifting towards machine-learned approaches for motion planning~\cite{autonomy_20}, the performance of self-driving systems is starting to rely heavily on large quantities of expert driving demonstrations. However, collecting this demonstration data typically involves expensive HD sensor suites (LiDAR + RADAR + cameras), which quickly becomes financially infeasible at the scales required. This motivates the use of commodity sensors like cameras for data collection, which are an order of magnitude cheaper than HD sensor suites, but offer lower fidelity. Leveraging these sensors for training an AV motion planner opens a financially viable path to observe the `long tail' of driving events.

As our main contribution we show it is possible to train a high-performance motion planner using commodity vision data which outperforms planners trained on HD-sensor data for a fraction of the cost. To the best of our knowledge, we are the first to demonstrate this using real-world data. We compare the performance of the autonomy system on these two different sensor configurations, and show that we can compensate for the lower sensor fidelity by means of increased quantity: a planner trained on 100h of commodity vision data outperforms the one with 25h of expensive HD data (see Fig.~\ref{fig:opener}). We also share the engineering challenges we had to tackle to make this work.
\end{abstract}

%%%%%%%%%%%%%%%%%%%%%%%%%%%%%%%%%%%%%%%%%%%%%%%%%%%%%%%%%%%%%%%%%%%%%%%%%%%%%%%%

\section{Introduction}

\begin{figure}[t!]
    \begin{center}
      \includegraphics[width=1\columnwidth]{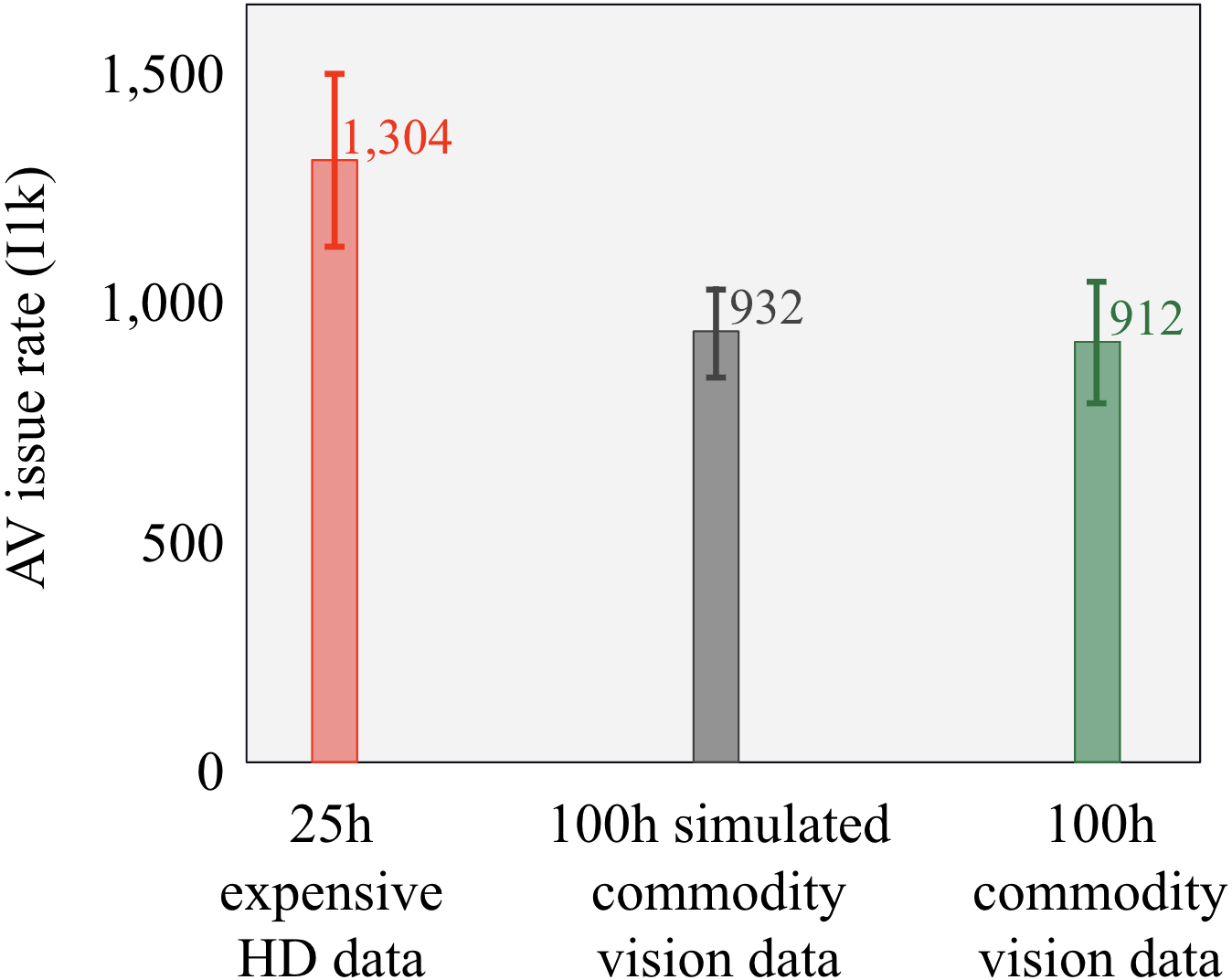}
    \end{center}
    
    \caption{\label{fig:opener} We compare the AV issue rate I1k (interventions per 1000 miles, lower is better) for the same motion planner architecture trained on three different datasets: 25h of expensive HD data (LiDAR, RADAR, cameras), 100h of simulated commodity vision data (HD data with reduced field-of-view and range), and 100h of commodity vision data (from real-world driving demonstrations). Training with 100h of commodity vision data (green) offers superior performance to 25h of expensive HD data (red). This indicates that we can achieve high-performance autonomy systems using large quantities of data that is an order of magnitude cheaper to collect, rendering ML approaches to motion planing \cite{autonomy_20} financially viable. We also note that our previous work \cite{chen2021data} on estimating of the value of commodity vision data via simulation (grey) was very close to reality (green). } 
\end{figure}

Autonomous driving promises a revolution in the human and goods transportation industries, but achieving human-level performance on the `long tail' of driving scenarios has proved extremely challenging. The typical bottleneck in performance occurs in the motion planner, which takes perception information as input and outputs a driving decision. The classical approach to motion planning has been rule-based, where engineers hand-design rules for every possible driving event. However, this approach scales poorly with the high-dimensional space of driving situations. This has motivated recent interest in imitation-learning~\cite{bansal2018chauffeurnet, gao_cvpr_2020, scheel2021urban, chen2021data}, i.e. using machine-learned (ML) models trained to mimic human behaviors from real-world driving examples. 

However, these approaches are extremely data-hungry, since the training data needs to capture enough of this long tail of driving situations. For example, \cite{bansal2018chauffeurnet} used millions of short driving demonstrations, while \cite{houston2020one} used tens of thousands of human driven miles.
This poses a new challenge, since these driving demonstrations are typically collected using fleets of vehicles equipped with expensive HD sensor suites: LiDARs, RADARs, and cameras \cite{bansal2018chauffeurnet, chen2021data}. Data collection at scale with these HD sensors is not a financially or operationally viable strategy. Using commodity sensors would make the approach viable, but these lower-fidelity data sources come at a cost: increased technical complexity. An ML planner relies on the accuracy of the perception system to learn to drive. The lower the sensor fidelity, the harder it is to achieve sufficiently accurate perception.

In this work we provide the first demonstration that it is possible to train an ML planner on real-world driving demonstrations collected with sensors that are much cheaper and lower fidelity, compared to the HD sensors used for operating the AV at test time. This is significant because it allows using regular non-autonomous vehicles equipped with commodity vision sensors to collect data at scale, making ML approaches to motion planning financially viable.

Specifically, we compare training the same ML planner on real-world data collected with two different sensors configurations: expensive HD sensors (LiDARs, RADARs, cameras) vs only commodity cameras. We perform an ablative analysis in controlled conditions, having equipped the same vehicles with both the expensive HD sensors and the cheaper commodity sensors.

This work offers the following contributions:
\begin{itemize}
    \item To the best of our knowledge, this is the first study showing it is possible to train an ML planner with data from sensors that are an order of magnitude lower-cost than those used at test time. Previous work analyzed this problem space in simulation~\cite{chen2021data}, i.e by adding noise to the accurate output of a perception system operating on HD sensors to make it look like the lower-fidelity output we would get from cheaper sensors. We compare these predictions to our real-world results (Fig.~\ref{fig:opener}).
     \item We demonstrate that training on 4x commodity vision data is better than on 1x expensive HD data.
    \item We demonstrate that the ML planner performance at test time improves with the amount of data used for training, even when we train it primarily on data collected with commodity sensors.
    \item We discuss the practical and technical challenges we had to face to make this work, such as reducing the domain gap between data collected with different sensors.
\end{itemize}

\section{Related work}

\subsection{Machine learning approaches to motion planning}
While classical approaches to motion planning heavily rely on rule-based expert systems (see~\cite{paden2016survey_classical} for an overview), there has been recent interest 
in machine-learning (ML) approaches trained on expert demonstrations~\cite{bansal2018chauffeurnet, gao_cvpr_2020, scheel2021urban, chen2021data}. This is based on the premises that: 1) ML models with a large number of parameters are better equipped to handle the significant cognitive complexity of a task like driving than rule-based approaches; and 2) performance improves with better data sets rather than by hand-engineering new driving rules~\cite{houston2020one, autonomy_20}.

Learnt approaches can be broadly categorized into end-to-end and mid-to-mid.
End-to-end methods consume raw sensor data and output steering commands, for example \cite{BojarskiTDFFGJM16} uses imitation learning for lane following, while~\cite{8957584} learns to drive end-to-end from simulation. We refer to \cite{survey_tampuu} for a broad review. Mid-to-mid planners (e.g. \cite{bansal2018chauffeurnet, scheel2021urban, chen2021data}) use an ML model that does not feed on the raw sensor data directly, but rather on the high-level 3D representation produced by intermediate sub-systems like localisation and perception. This intermediate representation makes it easier to interpret the final output. Recently, Zeng et al.~\cite{zeng2019end} proposed to bridge the two approaches with an end-to-end architecture producing (and trained on) intermediate outputs. 

In this paper we use the mid-to-mid approach. The intermediate representations used in mid-to-mid  (e.g. 3D trajectories and bounding boxes) decouple the perception and planning parts of the self-driving system and are typically sensor agnostic. This makes it possible to train the same ML planner architecture on driving demonstration collected with significantly different sensor configurations like we do here, and also allows comparing their performance more easily.

\subsection{Training ML planners on different data sources}
Using different data sources was previously explored through using synthetic data for training~\cite{muller2018driving} or validating~\cite{wong2020testing} a planner. Chen et al~\cite{chen2021data} simulated combining data from different sensor configurations by adding increasing levels of noise to real-world data \emph{entirely} collected with HD sensors. To the best of our knowledge, ours is the first attempt to combine real-world data collected with different sensor configurations to train a ML planner.

\subsection{Transfer learning and domain adaptation}
While the input format of the mid-to-mid planner is the same regardless of the sensor configuration, the properties of the perception data generated from the different sensor can be significantly different. Different sensor fidelity provides different levels of perception accuracy, and combining data collected with different sensors effectively changes the distribution of the inputs to the ML planner. Several techniques exist to handle domain shift between training and testing distribution (e.g.~\cite{5288526}), but little work has been done in the domain of motion planning~\cite{chen2021data}. One option is finetuning, which is widely used in literature, e.g.~\cite{vqa, antol2015vqa}. Another is domain adaptation, i.e. aiming to explicitly reduce the domain gap between domains, often used in computer vision, for example for synthesizing examples in new domains or style transfer \cite{8099501, hoffman2018cycada, li2018learning}. 

In our work, we focus on finetuning to address domain shift. Moreover, we leverage the recent progress in vision-only perception algorithms (see below), which mitigate the domain shift by reducing the perception accuracy gap between low and high fidelity sensors. We refer to~\cite{philion2020learning} for a study of the impact of perception accuracy on motion planning performance.

\subsection{Camera-based 3D perception}

A 3D perception system generates the intermediate representation that is provided as an input to the mid-to-mid ML planner at both train and test time. The system outputs 3D bounding boxes of objects and traffic agents, and we evaluate it in terms of their 3D accuracy~\cite{kitti}. Perception from multi-modal sensor suites that include expensive HD LiDARs achieve highest performance on traditional benchmarks~\cite{kitti, nuscenes2019}.  
Camera-based perception produces lower accuracy, with stereo camera systems (used in this paper) performing better than monocular solutions. However, for both monocular and stereo solutions the gap between vision-only sensor suites and LiDAR-equipped ones has reduced significantly over the last few years.

Camera-based perception methods can be grouped in three categories: direct~\cite{stereocnn}, voxel grid-based and Pseudo LiDAR-based~\cite{wang2019pseudo,pseudoLiDAR++}. We refer readers to~\cite{gsf-2021} for a detailed literature review. In this work, we use a voxel grid-based approach for our perception system for stereo camera data (Sec.~\ref{sec:data}). Grid-based methods discretize the world into a grid of 3D voxels. A neural network extracts 2D Image features and assigns them to 3D voxels via raycasting; this representation is then converted into more compact 2D bird's-eye-view (BEV) features for efficient downstream processing. While variants of voxel-grid based methods exist also to process monocular camera data~\cite{reading2021categorical}, the system we use is built on methods specialised for stereo inputs~\cite{chen2020dsgn}. In contrast to monocular methods, stereo inputs provide geometric constraints to localize objects in 3D space with better accuracy. Our 3D object detection framework draws inspiration from DSGN~\cite{chen2020dsgn}, as implicit voxelised representations outperform both direct and pseudo-LiDAR methods~\cite{liga-stereo}. DSGN builds a plane sweep volume (PSV) with pre-defined depth planes. It warps image-centric PSV features to 3D geometric space, projects them to birds-eye-view to perform 3D object detection in \etoe~manner. PSV is also used to estimate per pixel depth as an auxiliary task. We build upon this baseline model for 3D agent detection.

\begin{figure*}[thpb]
%\begin{figure*}[thpb]
    \vspace{2mm}
    \begin{center}
        \includegraphics[width=0.8\textwidth]{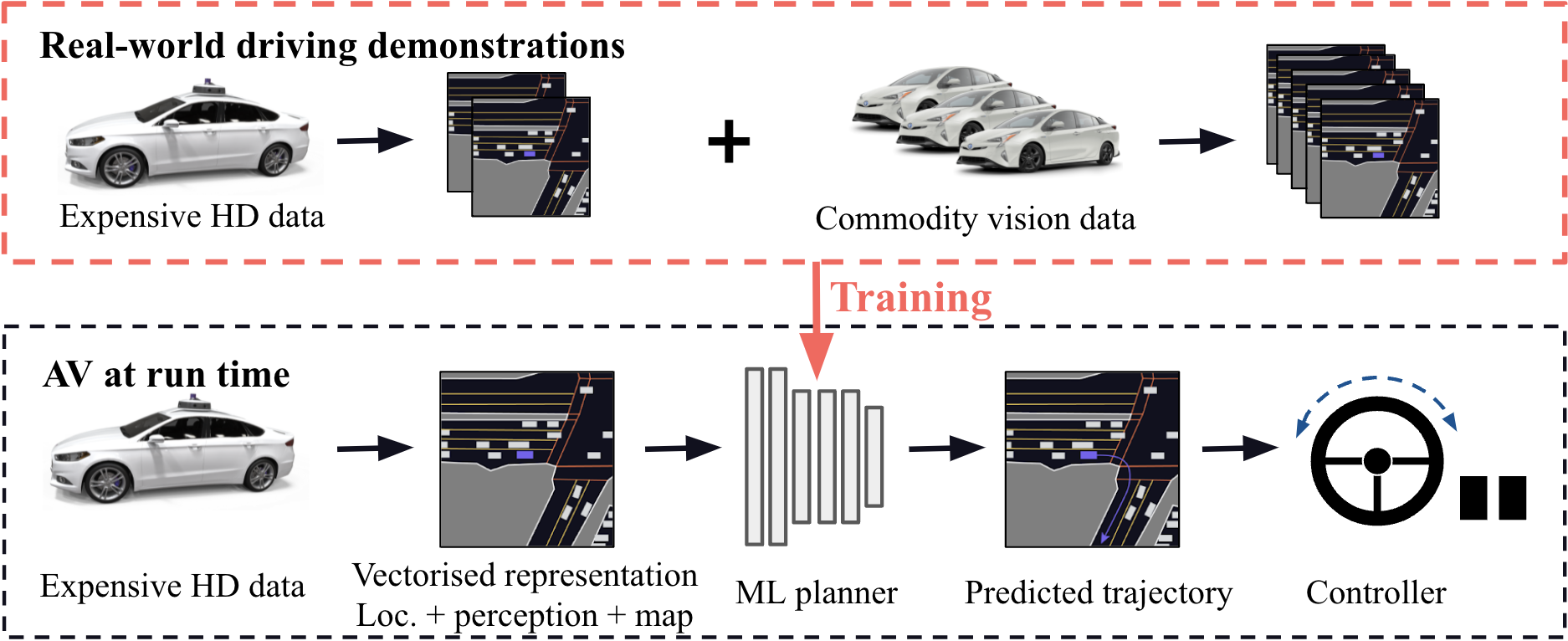}
    \end{center}
      \caption{\emph{Overview of the AV stack used in this work.} Bottom: At run-time, we use a traditional AV stack with a mid-to-mid planner (top, Sec.~\ref{sec:overview}). A 3D vectorised representation is extracted from the raw sensors data by combining the output of localisation, perception, and an HD map of the area. We feed this to the ML planner, which predicts a trajectory for the AV to follow that is then converted into control signals. Top: we train the ML planner on sensor data collected from vehicles with commodity sensor setup and fine tune it on HD data collected with expensive sensor suite. Thanks to the intermediate vectorised representation, we can train the planner on data collected with a variety of sensor configurations that can be different from those use at test time.}
      \label{fig:overview}
   \end{figure*}

\section{Methodology}

\subsection{Overview}
\label{sec:overview}
We use a traditional mid-to-mid AV stack (Fig.~\ref{fig:overview}, bottom) \cite{bansal2018chauffeurnet, scheel2021urban, chen2020simple, gao_cvpr_2020} and compare its performance as we train on data collected with different sensor configurations. When the planner is deployed onto an AV, we first extract a 3D vectorised representation from the raw sensor data. This combines the output of the localisation sub-system (i.e. 3D the position of the AV), the perception sub-system (the 3D poses of other traffic agents and obstacles), and a high-fidelity HD map of the area (like in~\cite{scheel2021urban}). This representation is the input to the ML planner, i.e. a neural network that predicts the trajectory for the AV to follow~\cite{bansal2018chauffeurnet, scheel2021urban, chen2020simple, gao_cvpr_2020}. The output of the network is then converted into the steering, throttle, and brake signals that control the vehicle.

We train the ML planner on real-world expert demonstrations (Fig.~\ref{fig:overview}, top). The sensor data is first converted into the same 3D vectorised representation used by the ML planner during run time. The supervision is provided by the recorded trajectory that was executed by the human driver during data collection~\cite{bansal2018chauffeurnet, scheel2021urban, chen2020simple, gao_cvpr_2020}. 

Thanks to the abstraction provided by the intermediate vectorised representation we can combine data from different sensor configurations during the training and the run-time of the ML planner. As we show in this paper, the sensors used to collect the training data can be cheaper and lower fidelity than those used by the AV at run-time. 

In Sec.~\ref{sec:planner} we provide more details on the ML planner we use in this work and how we train it, and then discuss inference (Sec.~\ref{sec:inference}) and performance evaluation (Sec.~\ref{sec:evaluation}). The methodology we introduce is generic; it enables the comparison of a variety of different sensor configurations (or even synthetic data~\cite{dosovitskiy2017carla,chen2021data,muller2018driving}), as well as different perception algorithms and HD map formats. The specific sensor modalities and perception algorithms we use in our experiments are presented in Sec.~\ref{sec:data}.

\subsection{ML planner}
 
\label{sec:planner}
The input to our ML planner is a vectorised representation encoding the state $\mathrm{s_t}$ of the environment around the vehicle at time $\mathrm{t}$. It contains bounding boxes for dynamic agents, state of traffic lights,  semantic map information like the position of lane boundaries and crosswalks (see Fig. \ref{fig:good-examples} for some visualisations of the input data). It is centered at the current position of the AV $\mathrm{x_t}$. We use the same input representation as in~\cite{scheel2021urban}, to which we refer for the details.
We also use the same transformer-based network architecture and backbone as the second baseline in ~\cite{scheel2021urban}, i.e. the behavior cloning model with perturbations. Please follow~\cite{l5kit} for implementation details.

Given the current state of the environment $\mathrm{s_t}$ and the recent history of the past $\mathrm{N}$ states, the network learns a function $\mathrm{f}$ that outputs the trajectory $\mathrm{X_t}$ for the vehicle to follow, i.e.  
$$
\mathrm{X_t =f(s_t,s_{t-N})}
$$
where $\mathrm{N=30}$. We represent $\mathrm{X_t}$ as a sequence of future positions
with a short time horizon of 1.2 seconds: $\mathrm{X_t = (x_{t+1}, \ldots, x_{t+T})}$ ($\mathrm{T=12}$ with time
increments of $\mathrm{0.1}$ seconds). This is using the same values as the experiments in \cite{scheel2021urban}.

We train the planner on sensor data collected by human drivers. For each timestamp in this
data, we first generate the vectorised representation to feed as input to the network. We then use the L2 distance between the trajectory predicted by the network and the trajectory $\hat{X}_t$ chosen by the human driver at that timestamp as training loss:
$$
    \mathrm{\mathcal{L} = \sum_{t=1}^T\norm{X_t - \hat{X}_t}}.
$$
 
The network predicts the trajectory for the next 1.2s ($\mathrm{T}$ = 12 steps). Following ~\cite{scheel2021urban,chen2021data}, we perturb the positions of the AV in the training and use Ackerman steering for realistic kinematics. The perturbations are needed to achieve better generalisation of the model for closed-loop testing~\cite{bansal2018chauffeurnet} (this is referred to as ``Behavioral Cloning + Perturbations'' in ~\cite{scheel2021urban}). Please note that for each timestamp in the training data, we always initialise the AV position to the current position of the human driver. This approach is referred to as \emph{open-loop} training.

\subsection{Inference}
\label{sec:inference}

During inference we run the ML planner model in \emph{closed-loop}. Specifically, at time $\mathrm{t}$, the vehicle predicts the trajectory $\mathrm{X_t}$. This prediction determines the position at time $\mathrm{t+1}$, which in turn is used to generate the input $\mathrm{s_{t+1}}$ that the model uses to predict $\mathrm{X_{(t+1)}}$. This is often referred to as unrolling. 
This is different to training, where we saw that for each time step we initialise the planner to the position of the human driver in the driving demonstration (Sec.~\ref{sec:planner}).

\subsection{Evaluation}
\label{sec:evaluation}
For our experiments, we follow~\cite{scheel2021urban} and run the ML planner in closed-loop simulation against a dataset of driving demonstrations (this is often referred to as log-based simulation). Simulation allows us to experiment in a more reproducible manner compared to road testing: we can compare different versions of the ML planner (i.e. trained on different datasets) by testing them on the exact same test set of driving demonstrations. 

A drawback of log-based simulation is that the behaviour chosen by the planner does not influence the rest of the environment, since we replay the motion of the traffic agents as recorded in the driving demonstration~\cite{scheel2021urban}, i.e. the traffic agents are non-reactive. Instead, in road-testing the state of the environment $\mathrm{s_t}$ is influenced by the decision taken by the planner at the previous step (e.g. the traffic agents behind the AV slow down if the AV slows down), and this is reflected in the perception output that is input to the planner. To limit the amount of drift the simulated AV can accumulate against the recorded positions, we run log-based simulation over relatively short time periods (30s) of a driving demonstration. 

We compute the same evaluation metrics as in~\cite{scheel2021urban}: the L2 distance to the human driver positions in the driving demonstration, the number of times the AV steers off the drivable surface or collides, and the overall I1k, i.e. the accumulated safety-critical interventions over 1000 miles. 
The test set we use to compute these metrics only contains data collected with HD sensors,
while we use different types of sensor data to train the ML planner, which we discuss next.

\subsection{Training data: expensive HD vs commodity vision} 
\label{sec:data}
\begin{figure}[t]
\begin{center}
   \includegraphics[width=0.8\linewidth]{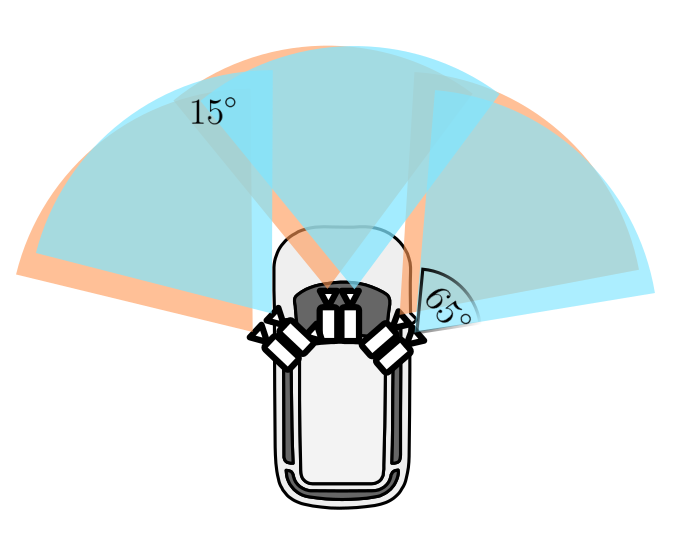}
\end{center}
   \caption{\emph{Commodity vision data}. We use a sensor suite consisting of only stereo cameras that is an order of magnitude cheaper than a typical HD sensor suite with HD LiDAR, RADARs, cameras, etc. We use 3 wide baseline stereo pairs that together provide a 160$^\circ$ field-of-view. Stereo cameras are lower-cost than HD, but allow to achieve good perception accuracy compared to even cheaper sensors like monocular cameras (Sec.~\ref{section:av_fleet}) }
\label{fig:stereo_setup}
\label{fig:onecol}
\end{figure}

We collect training data for the planner with two different sensor configurations: \textbf{expensive HD} and \textbf{commodity vision}. The HD data comes from a standard AV equipped with high-definition (HD) global-shutter cameras, state-of-the-art RADARs and HD LiDARs. For the commodity vision sensors, we use a setup consisting of 3 wide-baseline synchronised stereo cameras as shown in Fig.~\ref{fig:stereo_setup}, covering a field-of-view of around 160$^\circ$. The stereo sensors are roughly an order of magnitude less expensive than the HD AV sensors, but allow us to develop more accurate perception algorithms than even lower-cost sensors like monocular cameras.

We generate the vectorised representation to feed the ML planner (Sec.~\ref{sec:planner}) using different localisation and perception algorithms that are targeted to the two different sensor configurations. For HD sensors, we use proprietary state-of-the-art AV localization and perception algorithms which we consider to be industry-standard quality (an example can be found in \cite{lyft2019}). This data provides us with a full 360$^\circ$ field of view and range well over 100 meters. For stereo vision data, we use a visual mapping and localisation system~\cite{platinsky2020ar}, and a stereo detector dynamic object perception system~\cite{gsf-2021}. We build upon DSGN~\cite{chen2020dsgn} for the 3D object detection task. We adapt DSGN for efficient processing by optimizing depth planes, voxel resolutions, and voxel feature aggregation techniques. We use an anchor-free detector head~\cite{afdet} to predict 3D object attributes in contrast to the anchor-based approach in~\cite{chen2020dsgn}, to make the model computationally efficient. We achieve good-accuracy perception up to approximately 60 meters. We refer interested readers to~\cite{gsf-2021} for detailed evaluation and ablation studies of our HD and stereo vision perception methods.

For data collection, we mounted both the commodity and the HD sensors on the same vehicles. This allows us to observe the same driving experience with these two different sensor configurations and experiment in controlled conditions (the observed driving data is the same, the only thing changing is the sensor type and, as a consequence, the perception accuracy).

\begin{figure}[t]
    \centering
    \includegraphics[width=1.0\linewidth]{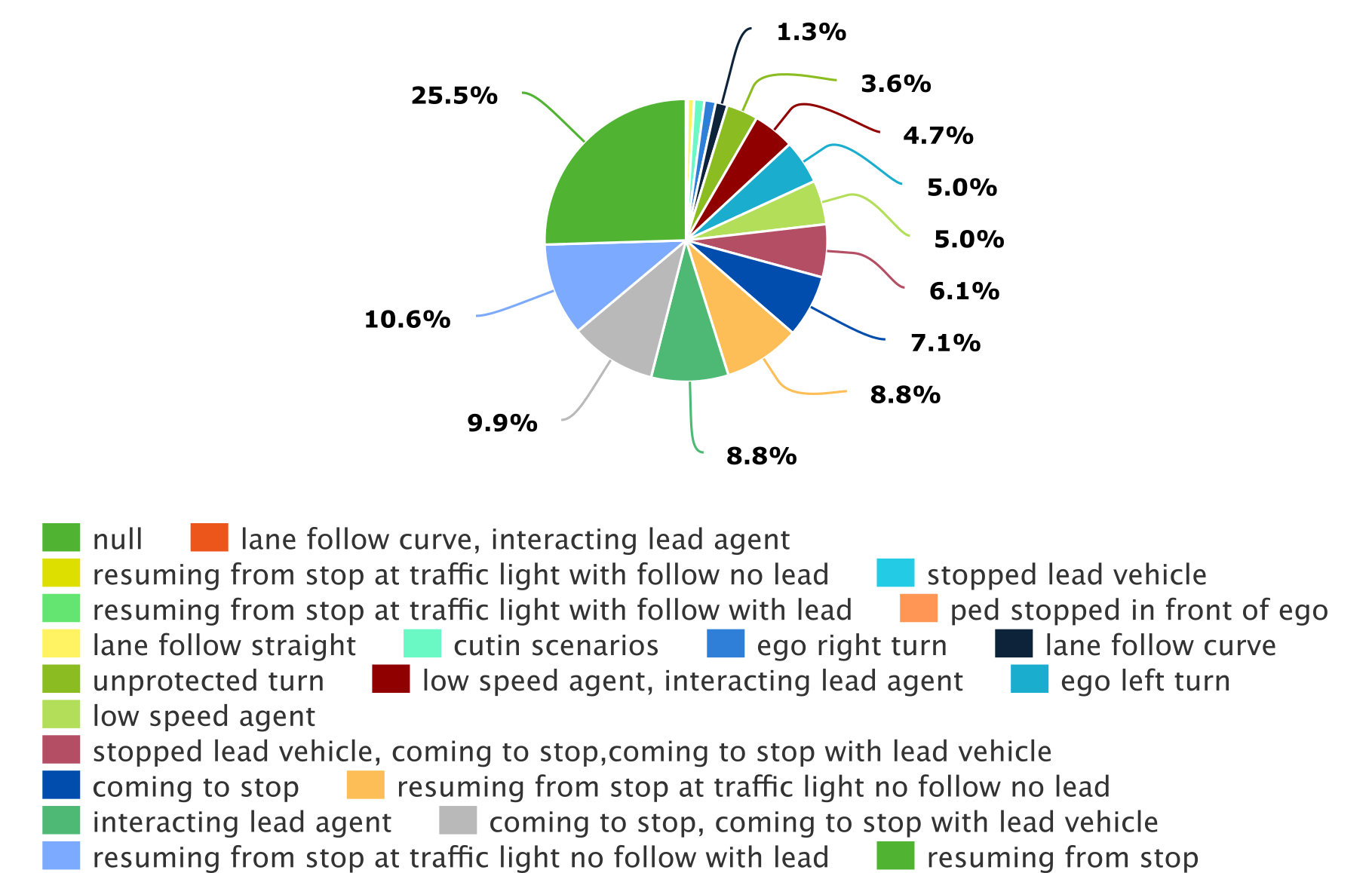}
    \caption{The data we use in our experiments (Sec.~\ref{sec:evaluation}) contains 140 hours of driving data (100 for training, 10 for finetuning, 30 for testing). This data is challenging and diverse, as we can see by the frequency of driving scenarios it contains.}
    \label{fig:scenario}
\end{figure}

We collected data focused on lane-follow behaviour over several months in downtown San Francisco, an urban driving environment with significant complexity. Overall, the dataset contains ~140 hours of 10 - 25 seconds long snippets - for each we have both the HD data and the commodity vision data version. We constructed this dataset from a larger set in order to include challenging and diverse driving scenarios as shown in Fig.~\ref{fig:scenario}. We use the following split: 100h for training, 10h for finetuning (to study the impact of domain shift, Sec.~\ref{section:domain_transfer}), and 30h of  for evaluation in simulation (Sec.~\ref{sec:evaluation}). When finetuning and evaluating, we always use the HD data version.

\subsection{Domain shift}
\label{section:domain_transfer}
% \subsubsection{Fine tuning and data mixing}
The vectorised representation $\mathrm{s_t}$ we input to the planner has different levels of accuracy and noise depending on the sensor used for collecting the data, as HD perception is more accurate than vision perception. This changes the distribution of the input between training and inference, introducing domain shift issues (see for example~\cite{chen2021data}). 
We use fine tuning to address this problem. We first train on commodity vision data and then do a few additional epochs on a small proportion of HD training data. 

\section{Experiments}

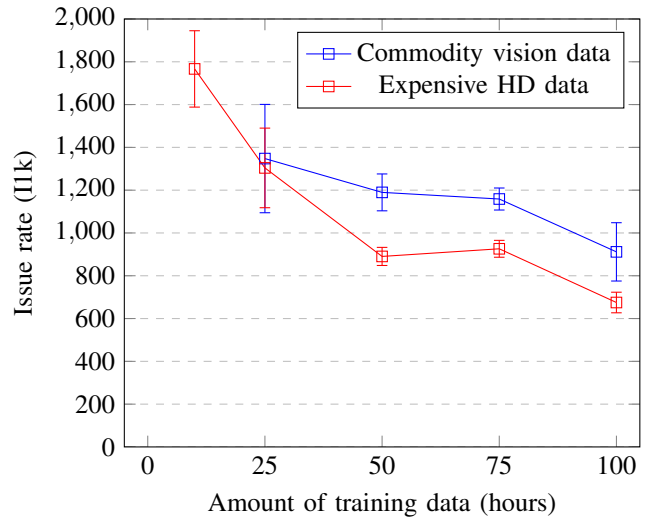
\begin{figure}[t]
\begin{tikzpicture}
\begin{axis}[
    xlabel={Amount of training data (hours)},
    ylabel={Issue rate (I1k)},
    xmin=-5, xmax=105,
    ymin=0, ymax=2000,
    xtick={0,25,50,75,100},
    ytick={0,200,400,600,800,1000,1200,1400,1600,1800,2000},
    legend pos=north east,
    ymajorgrids=true,
    grid style=dashed,
]

\addplot[
    color=blue,
    mark=square,
    ]
    plot [error bars/.cd, y dir = both, y explicit]
    coordinates {
    % (0,1766.666667) +- (0,178.442023)
    (25,1347.666667) +- (0,253.1433498)
    (50,1189.666667) +- (0,86.14845069)
    (75,1158.666667) +- (0,51.4414446)
    (100,911.6666667) +- (0,136.2750487)
    };
    
\addplot[
    color=red,
    mark=square,
    ]
    plot [error bars/.cd, y dir = both, y explicit]
    coordinates {
    (10,1766.666667) +- (0,178.442023)
    (25,1304) +- (0,186.023296)
    (50,890.3333333) +- (0,42.30313884)
    (75,926) +- (0,39.42080669)
    (100,675) +- (0,48.11098281)
    };

\legend{Commodity vision data,Expensive HD data}

\end{axis}
\end{tikzpicture}

\caption{Issue rate for the models (interventions per 1k miles, lower is better) when trained on increasing amounts of expensive HD and commodity vision data. We see that both models improve with more data. Importantly, we also observe that a model trained on 100h of commodity vision data outperforms the model trained on 25h of expensive HD data.}    
\label{fig:av_fleet}
\vspace{-6pt}
\end{figure}

\subsection{Comparing HD and commodity vision training data}
\label{section:av_fleet}
% We use the same amount of training data: 100 hours. As explained in Sec.~\ref{fig:av_fleet}, these are the exact same driving experiences observed by both the vision and HD sensor configurations, which are both installed on each vehicle.  
% Our evaluation test set consists of 30 hours of HD data and we report I1k as the main metric (Sec.~\ref{sec:evaluation}). In this experiment, we use fine tuning to handle domain shift (Sec.~\ref{section:domain_transfer}).
%An ablative analysis of these domain adaptation techniques is discussed in Sec.~\ref{section:domain_transfer_res} \todo{fix}.

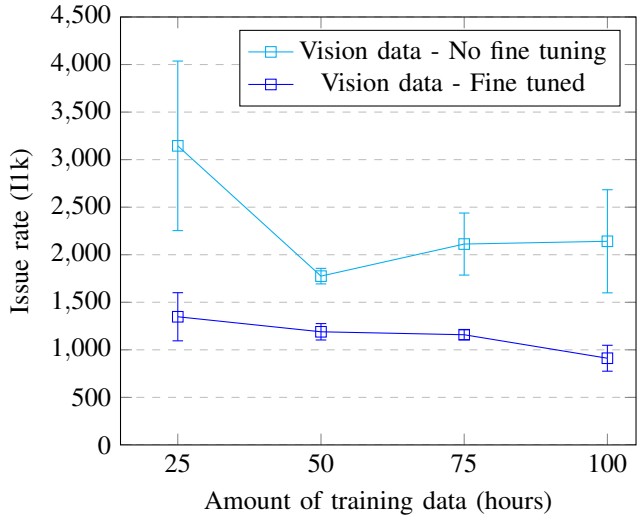
\begin{figure}[t]
\begin{tikzpicture}
\begin{axis}[
    xlabel={Amount of training data (hours)},
    ylabel={Issue rate (I1k)},
    xmin=15, xmax=105,
    ymin=0, ymax=4500,
    xtick={25,50,75,100},
    ytick={0,500,1000,1500,2000,2500,3000,3500,4000,4500},
    legend pos=north east,
    ymajorgrids=true,
    grid style=dashed,
]

\addplot[
    color=cyan,
    mark=square,
    ]
    plot [error bars/.cd, y dir = both, y explicit]
    coordinates {
    (25,3145) +- (0,891.6785669)
    (50,1775) +- (0,81.76796439)
    (75,2112.333333) +- (0,326.182294)
    (100,2141.333333) +- (0,542.1588533)
    };
    
\addplot[
    color=blue,
    mark=square,
    ]
    plot [error bars/.cd, y dir = both, y explicit]
    coordinates {
    (25,1347.666667) +- (0,253.1433498)
    (50,1189.666667) +- (0,86.14845069)
    (75,1158.666667) +- (0,51.4414446)
    (100,911.6666667) +- (0,136.2750487)
    };

% \addplot[
%     color=cyan,
%     mark=square,
%     ]
%     plot [error bars/.cd, y dir = both, y explicit]
%     coordinates {
%     (25,880) +- (0,0)
%     (50,524.5) +- (0,19.96820389)
%     (75,518.5) +- (0,19.96820389)
%     (100,655.5) +- (0,82.33303812)
%     };

\legend{Vision data - No fine tuning, Vision data - Fine tuned}%, Vision data - Mixed}

\end{axis}
\end{tikzpicture}

\caption{The domain gap between the 2 different sensor modalities (commodity vision data and expensive HD data). We can see that the model trained on commodity data requires fine tuning on HD data in order to perform well on HD data at inference time. This is due to different field-of-view, range, and noise properties of the 2 datasets.}    
\label{fig:finetune}
\vspace{-6pt}
\end{figure}

\begin{table*}[tbhp]
    \centering
    \def\arraystretch{1.4}
    \begin{tabular}{ @{}ccccccc@{}}
        \hline\toprule
        \multirow{2}{*}{\textbf{Training Data}} & \multicolumn{3}{ c }{\textbf{Collisions $\pm~(\sigma$)}} & \multicolumn{2}{ c }{\textbf{Imitation $\pm~(\sigma$)}}  & \multirow{2}{*}{\textbf{I1k} $\pm~(\sigma$)} \\
        & Front Collisions & Side Collisions & Rear Collisions & Off-road & L2 error & \\
        \hline
    
        10h HD (FT) & 37 $\pm$ (4) & 232 $\pm$ (11) & 306 $\pm$ (62) & 509 $\pm$ (7) & 1.36 $\pm$ (0.01) & 1767 $\pm$ (178) \\
        \hline
        25h vision & 29 $\pm$ (17) & 178 $\pm$ (30) & 761 $\pm$ (108) & 265 $\pm$ (82) & 0.89 $\pm$ (0.01) & 1348 $\pm$ (253) \\
        50h vision & 28 $\pm$ (9) & 160 $\pm$ (22) & 366 $\pm$ (50) & 255 $\pm$ (53) & 0.89 $\pm$ (0.03) & 1190 $\pm$ (86) \\
        75h vision & 19 $\pm$ (2) & 164 $\pm$ (23) & 739 $\pm$ (163) & 152 $\pm$ (37) & 0.85 $\pm$ (0.01) & 1159 $\pm$ (51) \\
        \textbf{100h vision} & \textbf{14 $\pm$ (5)} & \textbf{138 $\pm$ (7)} & \textbf{598 $\pm$ (116)} & \textbf{294 $\pm$ (39)} & \textbf{0.86 $\pm$ (0.03)} & \textbf{912 $\pm$ (136)} \\
        \hline
        \textcolor{blue}{\textbf{25h HD}} & \textcolor{blue}{\textbf{33 $\pm$ (4)}} & \textcolor{blue}{\textbf{171 $\pm$ (5)}} & \textcolor{blue}{\textbf{320 $\pm$ (21)}} & \textcolor{blue}{\textbf{481 $\pm$ (10)}} & \textcolor{blue}{\textbf{1.59 $\pm$ (0.01)}} & \textcolor{blue}{\textbf{1304 $\pm$ (186)}} \\
        50h HD & 24 $\pm$ (5) & 148 $\pm$ (1) & 356 $\pm$ (20) & 372 $\pm$ (22) & 1.21 $\pm$ (0.27) & 890 $\pm$ (42) \\
        75h HD & 22 $\pm$ (9) & 146 $\pm$ (14) & 435 $\pm$ (38) & 278 $\pm$ (18) & 1.25 $\pm$ (0.01) & 926 $\pm$ (39) \\
        100h HD & 8 $\pm$ (2) & 107 $\pm$ (6) & 514 $\pm$ (68) & 249 $\pm$ (45) & 0.95 $\pm$ (0.05) & 675 $\pm$ (48) \\
        \bottomrule\hline
    \end{tabular}
    \caption{\label{table:results}A breakdown of the metrics into: different collision types, ability of the algorithm to imitate the human driver and overall I1k. We observe that the model trained on 25 hours of expensive HD data is outperformed by that trained on 100 hours of commodity vision data and fine tuned on 10h HD data. For comparison we include the performance of the 10h HD fine tuning (FT) dataset. All metrics are averaged over 3 runs with standard deviation provided in the brackets.}
\end{table*}

We first measure the difference between training our ML planner on expensive HD data vs training it on commodity vision data. We compare performances for (25h, 50h, 75h, 100h) of commodity vision and expensive HD training data (Fig.~\ref{fig:av_fleet}). All models are finetuned on the 10h of HD data reserved for finetuning (for reference, we also include an ML planner trained on just the 10h finetuning data). We empirically observed that 38 epochs yielded the best results on the full 100h dataset. In each epoch we run inference and backpropagation on all the expert driving demonstrations available in the training dataset. As we decrease the amount of training data, we increase the number of epochs so that the models get approximately the same number of training steps to make the comparison fair (we use 152, 76, 50, and 38 epochs for the 25h, 50h, 75h, and 100h datasets respectively). Matching epochs instead of training steps resulted in a significantly worse performance due to shorter training when using small amounts of data (e.g. 25h). We also adapt the learning rate scheduling to match the training steps rather than epochs. For each data point, we compute error bars by aggregating results from three training jobs starting with differently initialised weights.

We highlight three takeaways.
First, performance increases with the amount of HD training data, which is
expected~\cite{houston2020one, chen2021data}: 100 hours performs twice as well as 10 hours. Second, we observe the same trend when training on commodity vision data, even if at test time we use HD data. Third, the decreased sensor data fidelity as a result of using commodity vision sensors can be overcome by increasing the \emph{quantity} of training data: 100 hours of vision data (912 I1k) outperforms 25 hours of HD training data (1,304 I1k).
These results confirm our hypothesis that it is possible to train a motion planner on data collected with sensors that are both cheaper and more scalable than those used at inference time. Our results with real-world data seem to be in line with previously simulated results by Chen et al.~\cite{chen2021data}.

\begin{figure*}[th!]
\begin{subfigure}{.16\textwidth}
  \centering
  \includegraphics[width=.95\linewidth]{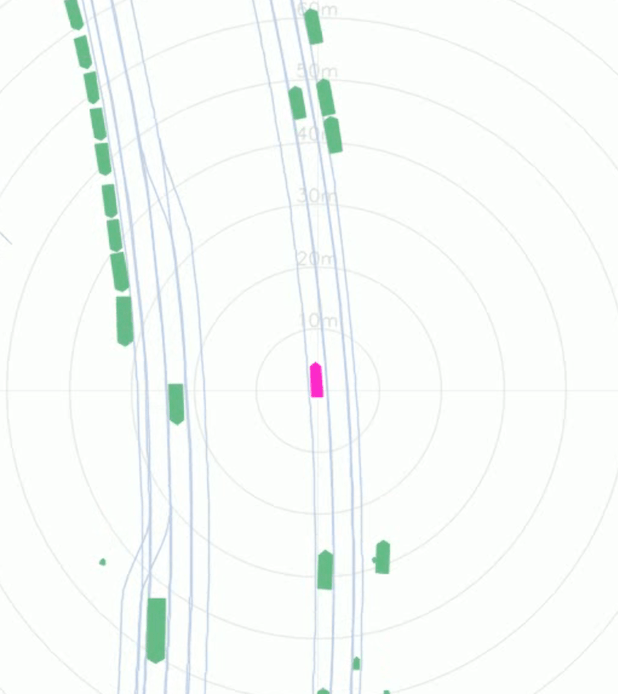}
%   \caption{1a}
  \label{fig:sfig1}
\end{subfigure}%
\begin{subfigure}{.16\textwidth}
  \centering
  \includegraphics[width=.95\linewidth]{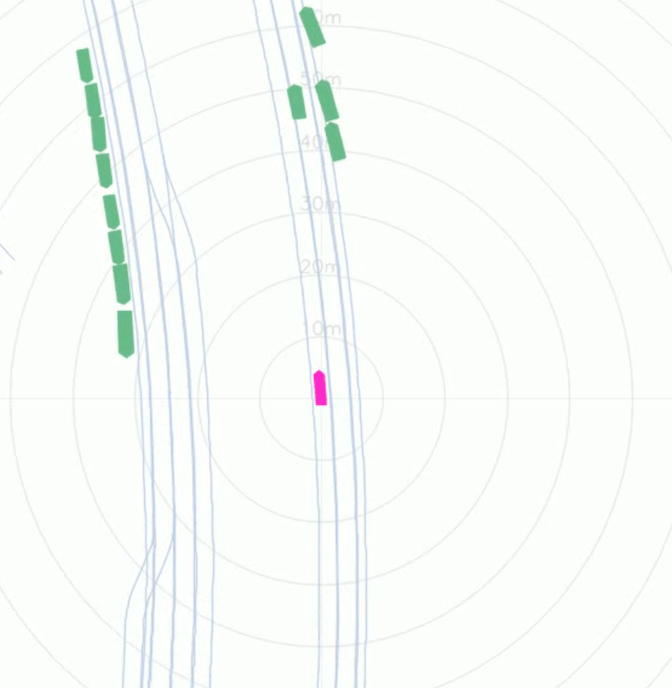}
%   \caption{1b}
  \label{fig:sfig2}
\end{subfigure}
\begin{subfigure}{.16\textwidth}
  \centering
  \includegraphics[width=.95\linewidth]{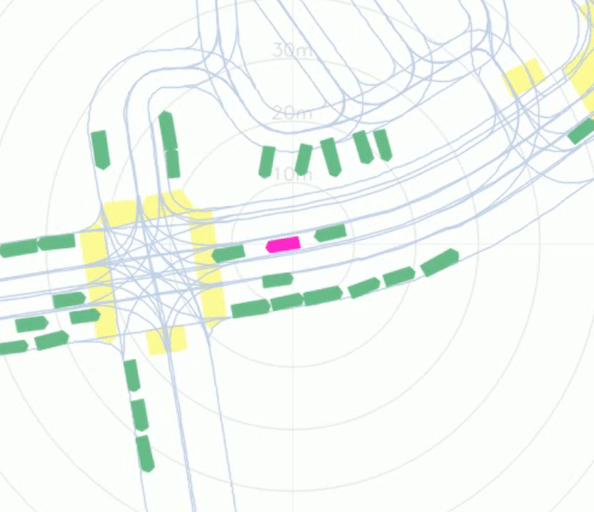}
%   \caption{1a}
  \label{fig:sfig1}
\end{subfigure}%
\begin{subfigure}{.16\textwidth}
  \centering
  \includegraphics[width=.95\linewidth]{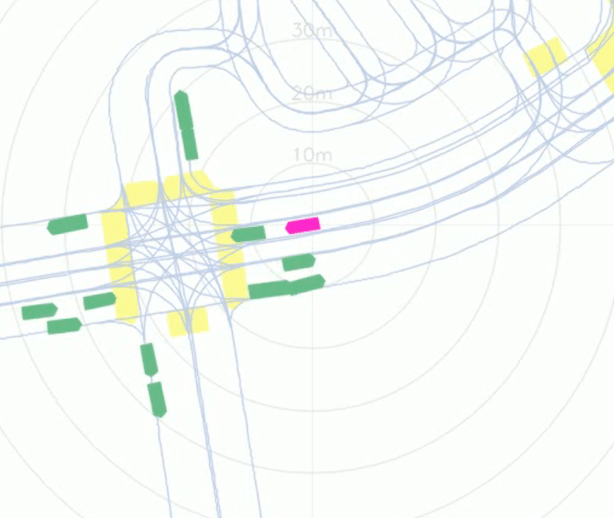}
%   \caption{1b}
  \label{fig:sfig2}
\end{subfigure}
\begin{subfigure}{.16\textwidth}
  \centering
  \includegraphics[width=.95\linewidth]{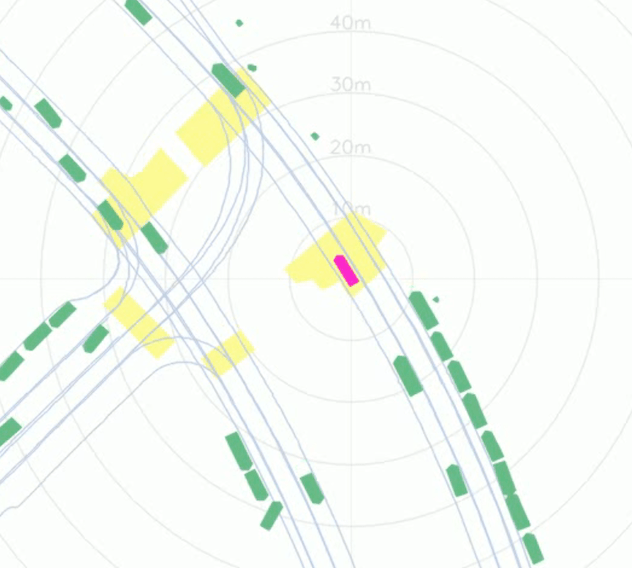}
%   \caption{1a}
  \label{fig:sfig1}
\end{subfigure}%
\begin{subfigure}{.16\textwidth}
  \centering
  \includegraphics[width=.95\linewidth]{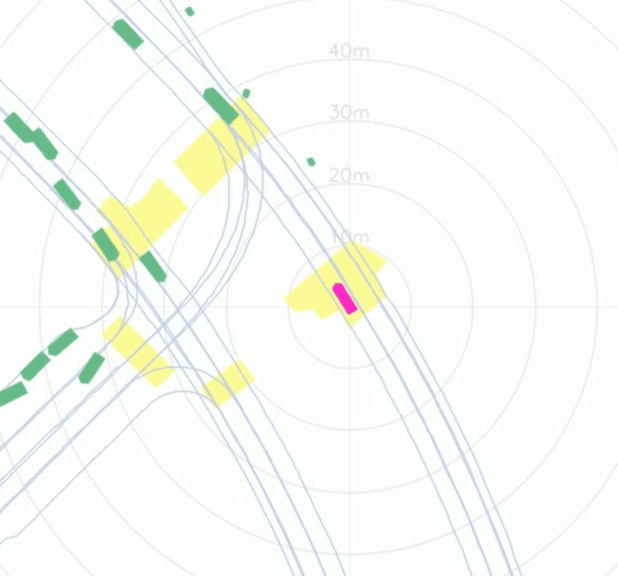}
%   \caption{1b}
  \label{fig:sfig2}
\end{subfigure}
\begin{subfigure}{.16\textwidth}
  \centering
  \includegraphics[width=.95\linewidth]{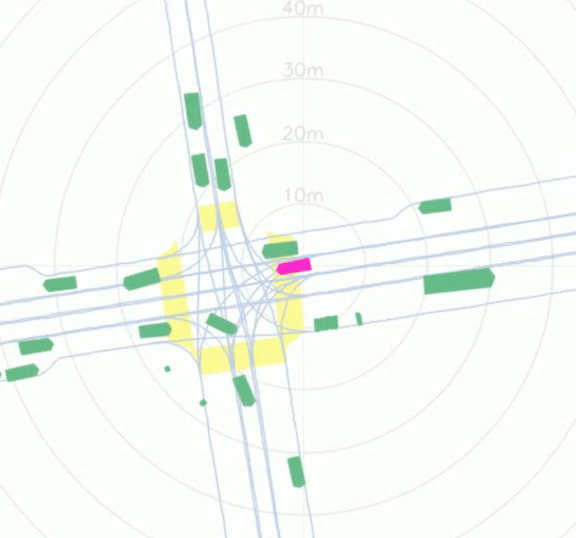}
%   \caption{1a}
  \label{fig:sfig1}
\end{subfigure}%
\begin{subfigure}{.16\textwidth}
  \centering
  \includegraphics[width=.95\linewidth]{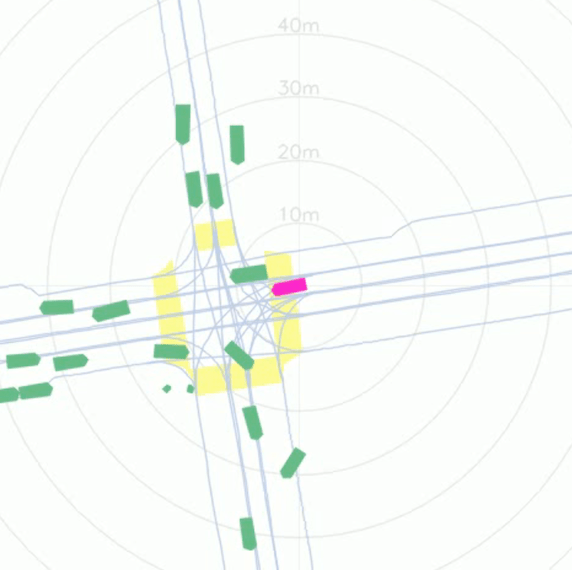}
%   \caption{1b}
  \label{fig:sfig2}
\end{subfigure}
\begin{subfigure}{.16\textwidth}
  \centering
  \includegraphics[width=.95\linewidth]{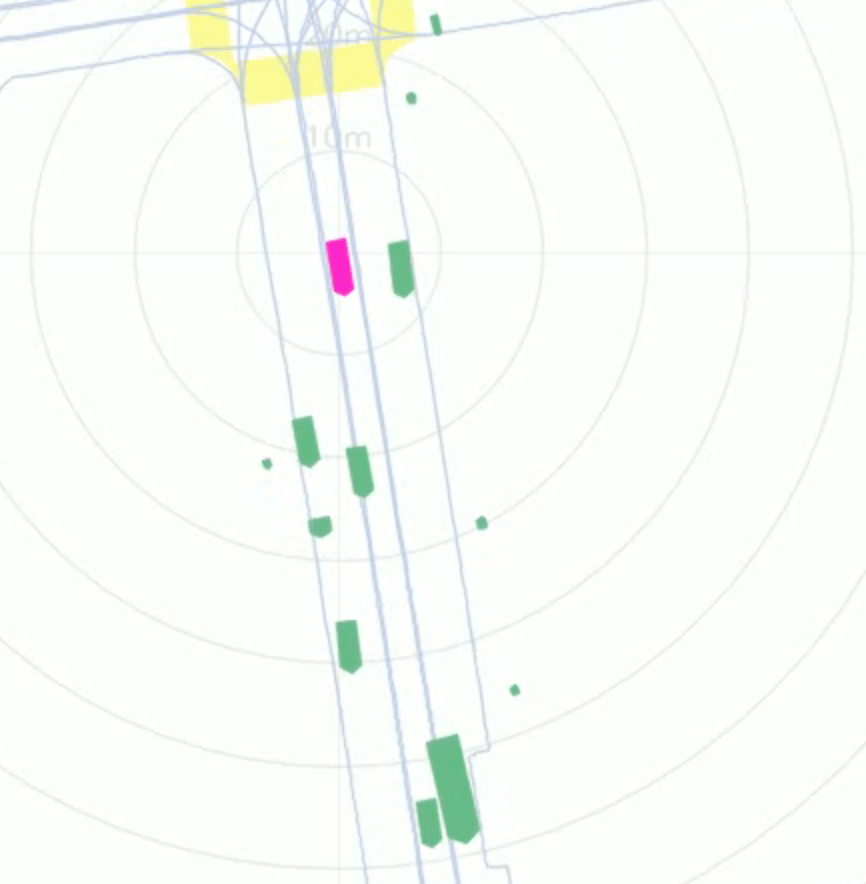}
%   \caption{1a}
  \label{fig:sfig1}
\end{subfigure}%
\begin{subfigure}{.16\textwidth}
  \centering
  \includegraphics[width=.95\linewidth]{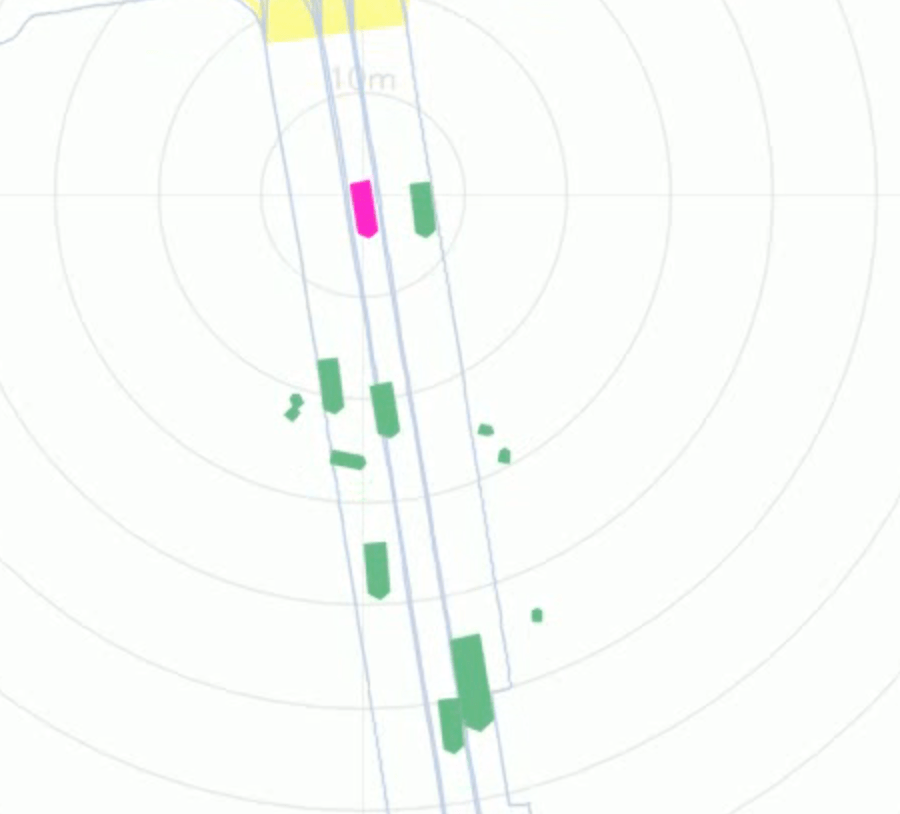}
%   \caption{1b}
  \label{fig:sfig2}
\end{subfigure}
\begin{subfigure}{.16\textwidth}
  \centering
  \includegraphics[width=.95\linewidth]{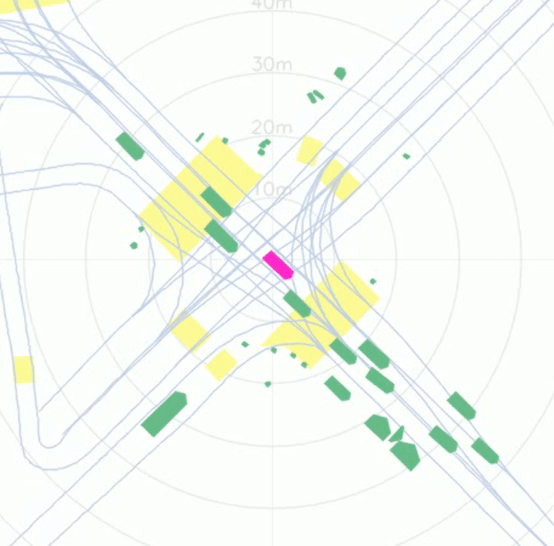}
%   \caption{1a}
  \label{fig:sfig1}
\end{subfigure}%
\begin{subfigure}{.16\textwidth}
  \centering
  \includegraphics[width=.95\linewidth]{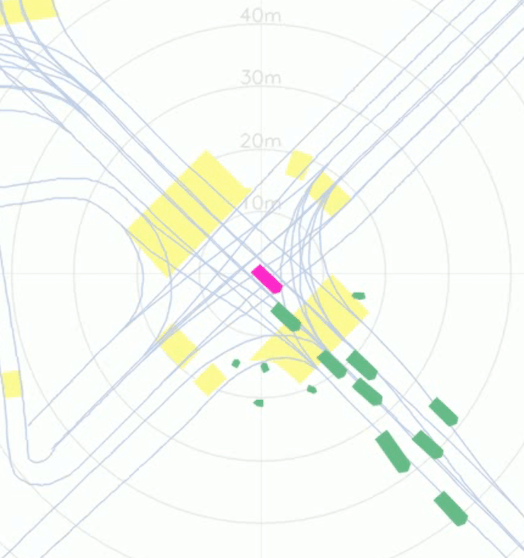}
%   \caption{1b}
  \label{fig:sfig2}
\end{subfigure}
\begin{subfigure}{.16\textwidth}
  \centering
  \includegraphics[width=.95\linewidth]{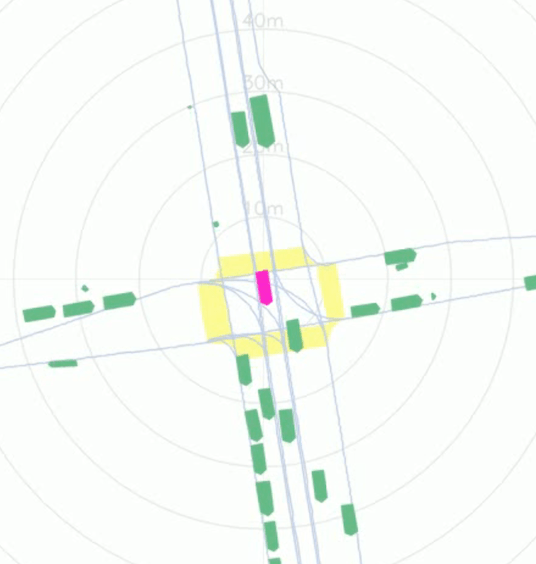}
%   \caption{1a}
  \label{fig:sfig1}
\end{subfigure}%
\begin{subfigure}{.16\textwidth}
  \centering
  \includegraphics[width=.95\linewidth]{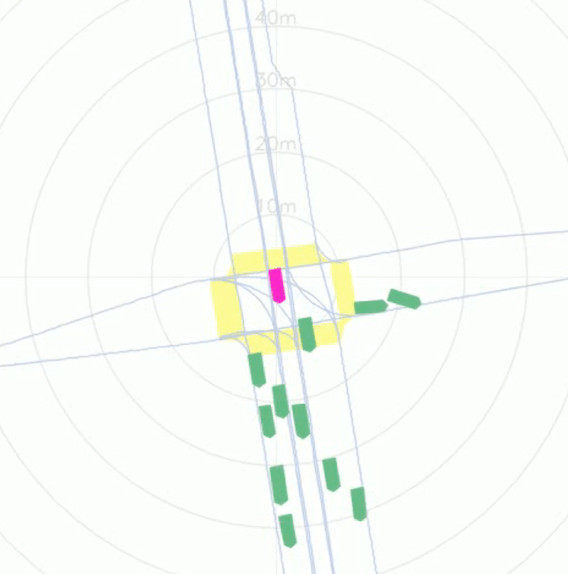}
%   \caption{1b}
  \label{fig:sfig2}
\end{subfigure}
\begin{subfigure}{.16\textwidth}
  \centering
  \includegraphics[width=.95\linewidth]{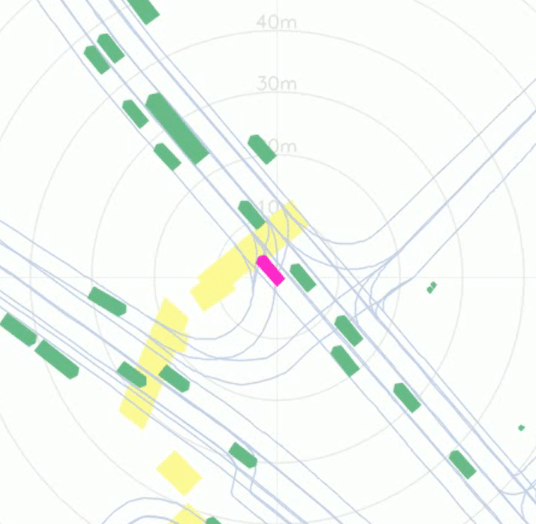}
%   \caption{1a}
  \label{fig:sfig1}
\end{subfigure}%
\begin{subfigure}{.16\textwidth}
  \centering
  \includegraphics[width=.95\linewidth]{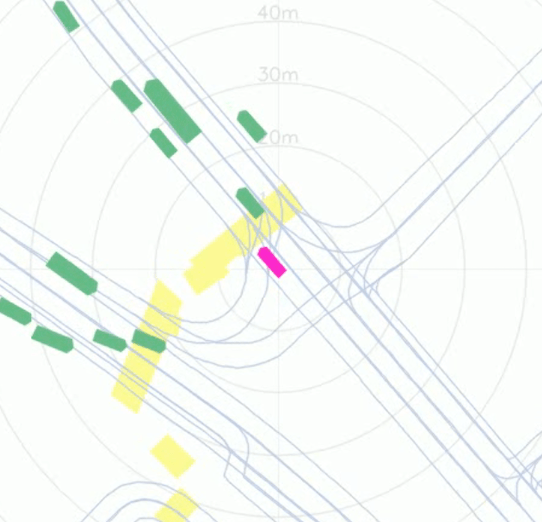}
%   \caption{1b}
  \label{fig:sfig2}
\end{subfigure}
\begin{subfigure}{.16\textwidth}
  \centering
  \includegraphics[width=.95\linewidth]{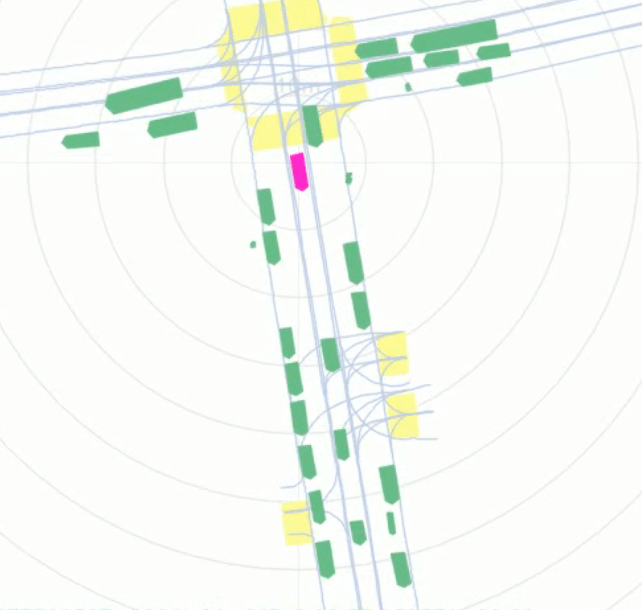}
%   \caption{1a}
  \label{fig:sfig1}
\end{subfigure}%
\begin{subfigure}{.16\textwidth}
  \centering
  \includegraphics[width=.95\linewidth]{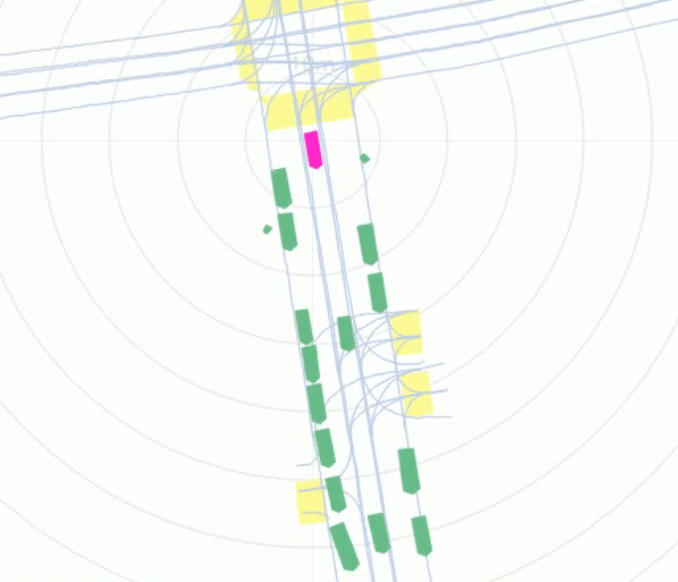}
%   \caption{1b}
  \label{fig:sfig2}
\end{subfigure}

\caption{Qualitative examples comparing the perception accuracy for expensive HD data (left of each pair) and commodity vision data (right). The cyan bounding boxes are detected agents (cars, pedestrians, etc). The pink bounding box is the AV. We also visualise traffic light states (by colouring the lane center they control in green or red), lanes and crosswalks. These examples show high precision and recall on agents within the field of view of the vision sensors. Best viewed in colour.}
\label{fig:good-examples}
\end{figure*}

Tab.~\ref{table:results} shows the performance of the different models broken down across safety, imitation performance, etc. The collisions are computed against all non-reactive agents present in the driving demonstration (see Fig. \ref{fig:good-examples} for some examples of the situations). We observe that the commodity-vision-trained models seem to be driving more cautiously and they suffer from more rear collisions with the non-reactive agents. As the evaluation dataset is focused on lane-follow scenarios (no lane changes are present) and we don't observe any increase in harsh breaking in our metrics, we do not expect the front-view-only vision system (Fig.~\ref{fig:stereo_setup}) to have an impact on rear collisions directly. We will explore using reactive agents to account for the rear collisions in future work. 

The main conclusion from comparing the 25h expensive HD and the 100h commodity vision results is that the vision model significantly outperforms the model trained on the smaller HD dataset in all metrics except for the rear collisions caused by non-reactive agents rather than a bad behaviour from the model. We note that vision-data models show more variance in the results, which indicates a need for future improvements to the model such that it can handle the commodity vision data with lower accuracy.

One interesting observation is that the L2 error of planners trained on HD data is noticeably larger than those trained on vision data. This is due to the fact that we terminate the fine tuning as soon as we see that the overall I1k error on the validation set starts to rise. For HD data models this happens early in the fine tuning procedure. With further fine tuning, L2 error keeps reducing, but we see significant regressions in I1k metrics. Hence, I1k performance and L2 imitation error do not always correlate. It is possible that the HD model starts to overfit to certain behaviours and finetuning does not provide the same generalization benefits as it does for the vision model.

\subsection{Fine Tuning}
\label{section:domain_transfer_res}
We perform an ablation study on the performance of the commodity vision models before and after fine tuning on the expensive HD data. We run fine tuning for five epochs using 1/10th of the learning rate. 
We stop fine tuning early if the I1k on the HD validation set stops improving. We evaluate on the same 30h HD dataset as in the previous experiment.

As expected, commodity vision data alone performs poorly due to domain shift (Fig.~\ref{fig:finetune}). Fine tuning the model on the HD data improves I1k by a factor of 2. We ran an additional experiment by fine tuning the models trained on HD datasets (the red curve in Fig.~\ref{fig:av_fleet}) on the same 10 hours of HD data used for fine-tuning to provide a fair comparison, but did not observe improvements in this case.

\subsection{Qualitative analysis of HD vs vision training data}

\begin{figure*}[th]
\begin{subfigure}{.16\textwidth}
  \centering
  \includegraphics[width=.95\linewidth]{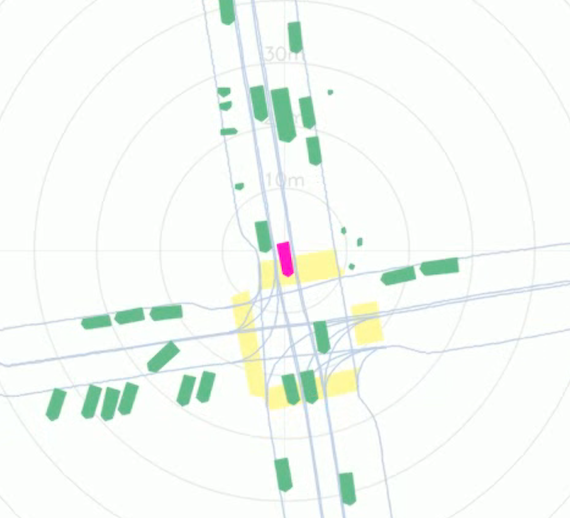}
%   \caption{1a}
  \label{fig:sfig1}
\end{subfigure}%
\begin{subfigure}{.16\textwidth}
  \centering
  \includegraphics[width=.95\linewidth]{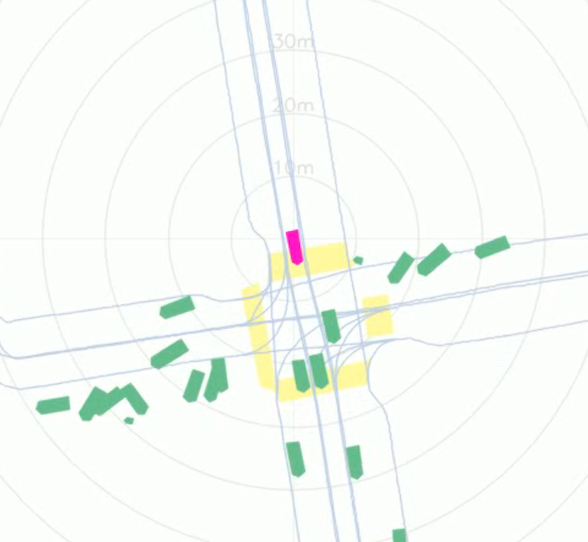}
%   \caption{1b}
  \label{fig:sfig2}
\end{subfigure}
\begin{subfigure}{.16\textwidth}
  \centering
  \includegraphics[width=.95\linewidth]{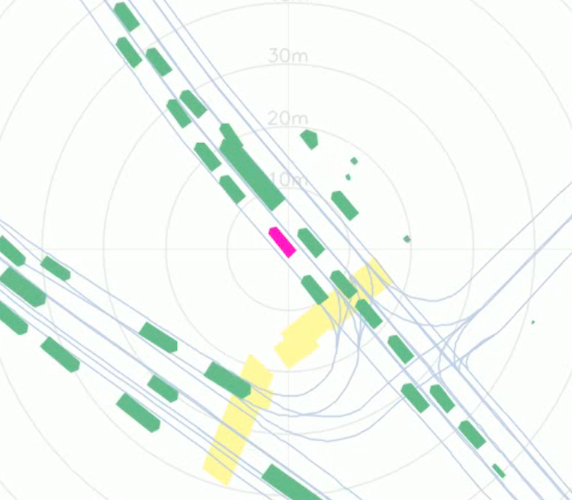}
%   \caption{1a}
  \label{fig:sfig1}
\end{subfigure}%
\begin{subfigure}{.16\textwidth}
  \centering
  \includegraphics[width=.95\linewidth]{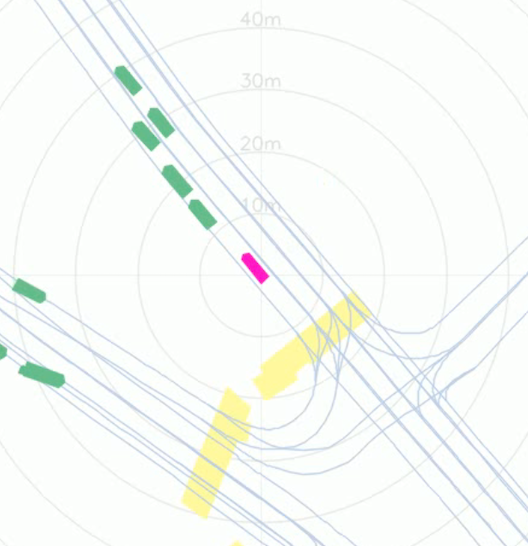}
%   \caption{1b}
  \label{fig:sfig2}
\end{subfigure}
\begin{subfigure}{.16\textwidth}
  \centering
  \includegraphics[width=.95\linewidth]{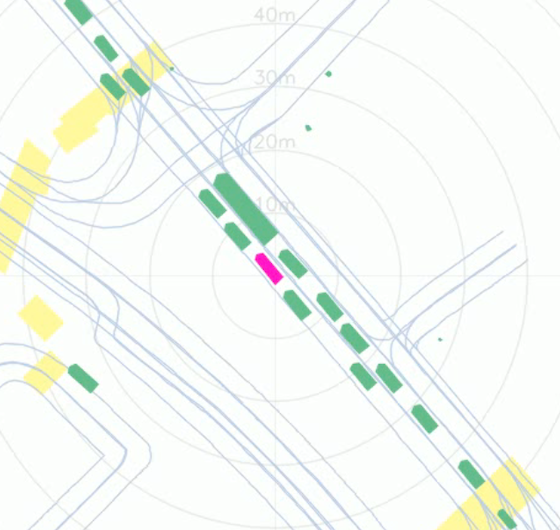}
%   \caption{1a}
  \label{fig:sfig1}
\end{subfigure}%
\begin{subfigure}{.16\textwidth}
  \centering
  \includegraphics[width=.95\linewidth]{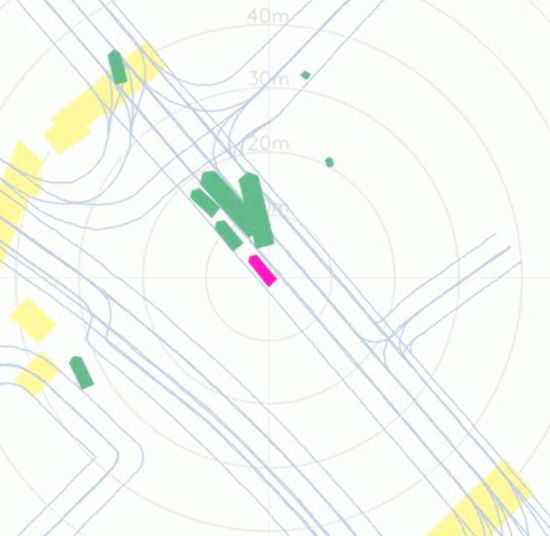}
%   \caption{1b}
%   \label{fig:bad-examples}
\end{subfigure}
\caption{Qualitative examples of failure cases, comparing the perception accuracy for expensive HD data (left of each pair) and commodity vision data (right of each pair). We observe noisy agents or missing agents in the vision data. Despite cases like these making their way into the training data, the ML planner can still leverage this data to improve its performance.}
\vspace{-12pt}
\label{fig:bad-examples}
\end{figure*}

The examples in Fig.~\ref{fig:good-examples} and \ref{fig:bad-examples} illustrate the differences in quality between the expensive HD data and the commodity vision data. The main difference is the lower field of view and range of the commodity vision sensors. Additional failure cases in the vision data (Fig.~\ref{fig:bad-examples}) include: missing agents, incorrect yaw prediction for agents, noisy detections (right-most example in Fig.~\ref{fig:bad-examples}), localization failures due to moving agents or changing lighting conditions.

We use a variety of methods for identifying failure cases: mining for training data containing traffic agents with unrealistic size, unrealistic ego motion, or large discrepancies between the ML planner prediction and the trajectory in the driving demonstration. Having the corresponding HD data available allows us to identify many failures of the vision-based algorithms, and we use it to triage and improve the results for the future. We estimate that $<10\%$ of the training data suffers from issues that can impact the planner.

\section{Conclusions}
Our results show that it is possible to train a motion planner on data collected with much cheaper sensors than those used at test time. We empirically show that we require 4x commodity vision data to bridge the performance gap of HD data, at $<$10x the HW cost (all other costs being approximately equal, e.g. data uploads, data processing and training of the ML planner). This is significant because it renders data collection at scale financially viable, by using fleets of regular (i.e. not autonomous) vehicles equipped with only cameras to collect large amounts of human driving demonstrations for training data-hungry ML planners. To the best of our knowledge, this is the first work that demonstrates this using real-world data. 

We also show that it is possible to address domain shift issues that inevitably arise with training and test data collected with different sensors using fine tuning. Our results demonstrate continual increase in performance as we add more data, implying that we can further close the performance gap between HD and commodity vision data given more training data, with an order of magnitude less cost.

In future work, we will study the generalisation of our findings along four important axes: (1) \emph{data volumes}: does performance keep improving as we scale amounts of training data to O(10K)+ hours of data? This will require developing strategies to mine interesting data (e.g. complex \& rare scenarios), so as to balance the training data and keep the computational requirements of model training manageable; (2) \emph{geography}: while the data used in this paper already includes a very diverse set of driving manoeuvres (Fig.~\ref{fig:scenario}) in a fairly wide area (San Francisco), we will analyse expanding to wider geographies (e.g. multiple US states); (3) \emph{ML planner backbones} (Sec.~\ref{sec:planner}); (4) \emph{sensor configurations}: while we already demonstrated generalisation to multiple sensor configurations (expensive HD and stereo), we will study even more commodity ones, like monocular cameras. We note that the ability to leverage commodity sensor data we demonstrated in this work is a key enabler to pursue these directions, as for example significant expansion of data volumes and geographies would be financially prohibitive if leveraging only HD sensor configurations.


%%%%%%%%%%%%%%%%%%%%%%%%%%%%%%%%%%%%%%%%%%%%%%%%%%%%%%%%%%%%%%%%%%%%%%%%%%%%%%%%

%%%%%%%%%%%%%%%%%%%%%%%%%%%%%%%%%%%%%%%%%%%%%%%%%%%%%%%%%%%%%%%%%%%%%%%%%%%%%%%%



% \section*{ACKNOWLEDGMENT}

% The preferred spelling of the word “acknowledgment” in America is without an “e” after the “g”. Avoid the stilted expression, “One of us (R. B. G.) thanks . . .”  Instead, try “R. B. G. thanks”. Put sponsor acknowledgments in the unnumbered footnote on the first page.

%%%%%%%%%%%%%%%%%%%%%%%%%%%%%%%%%%%%%%%%%%%%%%%%%%%%%%%%%%%%%%%%%%%%%%%%%%%%%%%%

\bibliographystyle{IEEEtran}
\bibliography{references}

\end{document}